# ВЫПУСКНАЯ КВАЛИФИКАЦИОННАЯ РАБОТА

**Исследовательский проект на тему**

**Оптимизация применения oblivious решающих деревьев на ЦП**

**Выполнил студент группы 186, 4 курса,
Миронов Алексей Дмитриевич**

**Руководитель ВКР:
Кандидат технических наук, доцент, Сухорослов Олег Викторович**

**Соруководитель ВКР:
Кандидат физико-математических наук, руководитель группы инфраструктуры качества поиска, Хузиев Ильнур Масхудович**

# Abstract

CatBoost is a popular machine learning library. CatBoost models are based on oblivious decision trees, making training and evaluation rapid. CatBoost has many applications, and some require low latency and high throughput evaluation. This paper investigates the possibilities for improving CatBoost's performance in single-core CPU computations. We explore the new features provided by the AVX instruction sets to optimize evaluation. We increase performance by 20–40% using AVX2 instructions without quality impact. We also introduce a new trade-off between speed and quality. Using float16 for leaf values and AVX-512 instructions, we achieve 50-70% speed-up.

**Keywords** – oblivious decision tree ensembles; CatBoost; MatrixNet; SIMD; SSE; AVX; AVX-512; FP16.

# Аннотация

CatBoost – популярная библиотека машинного обучения, использующая oblivious решающие деревья, обеспечивающие высокую скорость обучения и применения моделей. CatBoost используются во многих областях, в том числе в тех, где время применения моделей критично. В рамках этой статьи мы исследуем возможности оптимизации однопоточного применения CatBoost моделей на центральном процессоре с помощью AVX инструкций. Мы достигли ускорения на 20–40% с использованием AVX2 инструкций без влияния на качество, а также ускорения на 50–70% с использованием AVX-512 инструкций и уменьшением качества за счёт использования чисел с плавающей точкой половинной точности в листьях.

**Ключевые слова**: oblivious решающие деревья, CatBoost, MatrixNet, SIMD, SSE, AVX, AVX-512, FP16.

# 1. Оглавление

## 2. Введение

Лес решающих деревьев – популярный алгоритм машинного обучения. Способность решающих деревьев находить зависимости между разнотипными данными делает этот алгоритм незаменимым в ряде областей. В некоторых из них скорость применения моделей является критичной. Примером такой задачи может являться ранжирование документов поисковой системой. Решающие деревья используются многими поисковыми системами, включая Яндекс.Поиск и Microsoft Bing [2], и от этих систем требуется определить релевантность как можно большого числа документов за небольшой интервал времени. В таких крупных системах даже небольшая оптимизация может значительно сказаться на расходах на серверные мощности. Так же при более эффективном применении модели система может ранжировать больше документов, используя более тяжёлые модели, повышая качество поисковой выдачи.

В классическом виде дерево поиска представляет из себя подвешенное бинарное дерево, где в листьях содержатся значения целевой функции (ответы/предсказания), а во всех остальных вершинах (внутренних узлах) – условия, которые могут быть истины или ложны в зависимости от аргументов функции (признаков объектов). Чтобы получить значение по аргументам, нужно пройти по дереву, начиная в корне и на каждом шаге идя в левое или правое поддерево в зависимости от истинности условия текущей вершины, пока не окажешься в листе.

Быстрое применение классических решающих деревьев является нетривиальной задачей в том числе из-за проблемы, известной как control hazard.

При наивном обходе такого дерева, чтобы узнать следующую вершину, нужно вычислить условие в текущей. Это значит, что мы не можем поместить инструкции для следующей вершины в конвейер процессора пока не выполняться инструкции текущей, что означает простой конвейера. Так же это означает, что мы не сможем заранее подгружать нужные данные, и приходя в вершину нам придётся ждать данные из памяти.

Ещё одной проблемой наивного обхода является сложность векторизации. SIMD (Single Instruction Multiple Data) инструкции – набор инструкций процессора, позволяющий выполнять одну операцию для нескольких значений одновременно. SIMD инструкции позволяют распараллелить обработку данных. Мы можем распараллелить процесс, применяя деревья для нескольких объектов сразу, но в условиях control hazard это не даст значительного ускорения, так как перед выполнением инструкции, нужно будет ждать данных, лежащих в разных кэш-линиях.

Для борьбы с control hazard было предложено множество алгоритмов. Наиболее эффективным на данный момент скорее всего является RapidScorer [2]. Однако, в то время как наивный обход для каждого объекта вычисляет условия только в вершинах на пути от корня до листа с ответом, алгоритмы вроде Rapid Scorer вычисляют условия во всех вершинах дерева. То есть, эти алгоритмы работают быстрее наивного обхода за счёт векторизации и устранения control hazard, но при этом делают намного больше вычислений, большая часть из которых не нужна и не влияет на итоговый ответ.

Мы можем устранить control hazard и при этом не делать лишних операций используя другую структуру деревьев – oblivious деревья. Oblivious деревья имеют одинаковые условия в вершинах на одном уровне (на одном расстоянии от корня). То есть на любом пути от корня до листа встречается одинаковый набор условий. Даже наивный обход в таком случае лишён ветвлений вплоть до загрузки значения из листа.

Разумеется, oblivious деревья дают менее качественные предсказания, но, так как их применение более быстрое, мы можем увеличить число деревьев в модели, что может улучшить качество.

Среди библиотек, имплементирующих oblivious решающие деревья, наиболее популярной является CatBoost. CatBoost разрабатывается компанией Яндекс и имеет несколько тысяч отметок "Star" на GitHub [4]. CatBoost может применять модели как на центральном, так и на графическом процессоре. В рамках этой статьи мы постараемся ускорить применение CatBoost моделей на ядре центрального процессора с помощью SIMD инструкций.

Текущая реализация CatBoost написана на SSE3, появившемся на процессорах в 2004 году, и предоставляющем 16 128-битных регистров и 227 инструкций. Сейчас на многих процессорах уже доступно расширение AVX-512, предоставляющее 32 512-битных регистра и больше 5000 инструкций.

Оставшаяся часть статьи организована следующем образом. В 3-й главе описываются существующие решения, включая алгоритм CatBoost, который мы будем оптимизировать. 4-я глава посвящена деталям работы процессора, важным для эффективной реализации векторизации. В 5-й мы описываем методологию исследования. В главах 6–8 приводятся гипотезы для каждой из трёх этапов применения, описанных в третьей главе. В 9-й мы изучаем результаты замеров времени применения, и определяем, какие из гипотез оказались успешны. 10-я глава рассказывает об ограничениях предложенных нами алгоритмов и перспективах дальнейшей работы.

## 3. Существующие решения

Алгоритмы векторизации для классических (ассиметричных) деревьев, такие как V-QuickScorer [3] и RapidScorer [2], решают проблемы, которые даже не возникают при работе с oblivious деревьями, поэтому в рамках этой статьи рассматривать их мы не будем.

## 3.1. MatrixNet

MatrixNet – предок CatBoost. MatrixNet также работает с oblivious деревьями. В настоящее время в большинстве областей CatBoost заменил MatrixNet, однако последний до сих пор используются в том числе в поиске.

MatrixNet модель, принимает на вход массив 32-битных чисел с плавающей точкой $FEATURES$, содержащий признаки объекта, и возвращает одно число. При обучении модели для каждого числового признака выбирается несколько пороговых значений $THRESHOLDS$. При обучении и применении числовые признаки конвертируются в бинарные путём сравнения их значений с порогами. $BINFEATURES_i \coloneqq (FEATURES_{INDEX_i} < THRESHOLDS_i)$, где $BINFEATURES$ – бинарные признаки объекта, $i$ – индекс бинарного признака, а $INDEX$ – массив, задающий соответствие бинарного признака числовому. Этот процесс называется **бинаризация**. Каждый бинарный признак задаётся индексом числового признака и порогом.

В каждой промежуточной вершине содержится условие, задаваемое бинарным признаком. При обходе дерева мы выбираем следующую вершину в зависимости от значения бинарного признака текущей вершины. Один бинарный признак может использоваться несколькими деревьями.

Легко заметить, что число используемых oblivious деревом бинарных признаков равно высоте дерева. Более того, индекс листа с ответом в бинарной записи состоит из бит, соответствующих бинарным признакам. Например, если у нас есть дерево высоты 3, и в корне условие выполняется (значение бинарного признака равно 1), на расстояние один от корня не выполняется (0), и на расстоянии два выполняется (1), то обходя дерево мы придём в лист с индексом $101_2 = 5$. Это позволяет нам легко вычислить индексы листьев с нужными ответами по бинарным признакам с помощью битовых сдвигов и дизъюнкций.

Получив индексы, нужно загрузить соответствующие значения из листьев. В MatrixNet значения в листьях являются 32-битными целыми числами. Для получения финального результата предсказания из листьев конвертируются в 64-битные числа и суммируются по деревьям. Полученная сумма конвертируется в 64-битное число с плавающей точкой умножением на *scale* и прибавлением *bias*.

Для простоты можно разделить применение MatrixNet модели на три этапа:

- Бинаризация – превращение числовых признаков объектов в бинарные
- Вычисление индексов листьев – для каждого дерева загрузка нужных бинарных признаков и вычисление индексов битовыми сдвигами
- Загрузка значений листьев – загрузка значений из листьев и суммирование их по всем деревьям

Для эффективной векторизации объекты обрабатываются группами. Каждый из этих трёх этапов можно векторизовать разными способами. Про способы векторизации в MatrixNet мы проделали исследование в прошлом [10].

Для бинаризации можно использовать векторные инструкции для сравнения нескольких чисел с порогами за раз, при этом можно использовать регистры разного размера в зависимости от доступных расширений процессора. Так же SIMD инструкции можно использовать для лучшей утилизации шины процессора для эффективной работы с памятью. Полученные биты можно укладывать в память в разном порядке. Для вычисления индексов листьев всё аналогично.

С загрузкой значений из листьев сложнее. Можно загружать их по индексам без использования SIMD инструкций и использовать их только для суммирования этих ответов. Но эффективно утилизировать шину процессора для загрузки небольших по размеру чисел из разных участков памяти сложно. Альтернативой можно использовать SIMD инструкции семейства Gather или загрузить все ответы дерева в регистры и использовать permute инструкции, чтобы сопоставить индексам ответы.

### 3.2. CatBoost

CatBoost – популярная библиотека градиентного бустинга с открытым исходным кодом, созданная на основе MatrixNet. Во многих бенчмарках

CatBoost обходит по производительности другие популярные библиотеки за счёт использования oblivious деревьев. В отличии от MatrixNet CatBoost поддерживает и классические (ассиметричные) деревья, тем не менее оптимизация применения классических деревьев выходит за рамки данной работы.

Другим важным отличием является поддержка CatBoost категориальных признаков [1]. Категориальные признаки бинаризуются одним из двух способов: one-hot кодированием или ctr. Ctr – замена категориальных признаков на какие-то числовые статистики, посчитанные для этих признаков. После замены эти признаки бинаризуются точно так же, как числовые. One-hot как правило не используются CatBoost при большом числе категорий [9], поэтому его оптимизация не представляет интереса.

Алгоритм применения похож на MatrixNet, его можно также условно разделить на три этапа. Доля времени каждого из этапов от общего времени применения модели можно увидеть на рисунке 3.1.

В отличии от MatrixNet, в CatBoost бинаризация выполняется и записывается в сжатом формате. Для числовых и ctr признаков считаются квантили: для каждого значения считается число порогов, которые это значение переступает. Чтобы обернуть операцию – понять по квантилю, удовлетворяет ли значение конкретному порогу, достаточно сравнить квантиль с порядковым номером порога, при условии, что границы упорядочены. Квантили – однобайтовые. Если порогов больше 254, то они разделяются на группы размера до 254. В случае с one-hot кодированием, просто пишется номер категории, начиная с 1. Если категорий больше 254, то они точно также делятся на группы. Если значение не соответствует ни одной категории (вообще или в текущей группе), то пишется 0.

Так же CatBoost умеет решать задачи многоклассовой классификации. Для этого в листьях может писаться не одно значение, а несколько – по значению для каждого класса.

Текущая реализация применения написана с использованием SSE инструкций. Мы постараемся улучшить её, в том числе с помощью AVX инструкций.

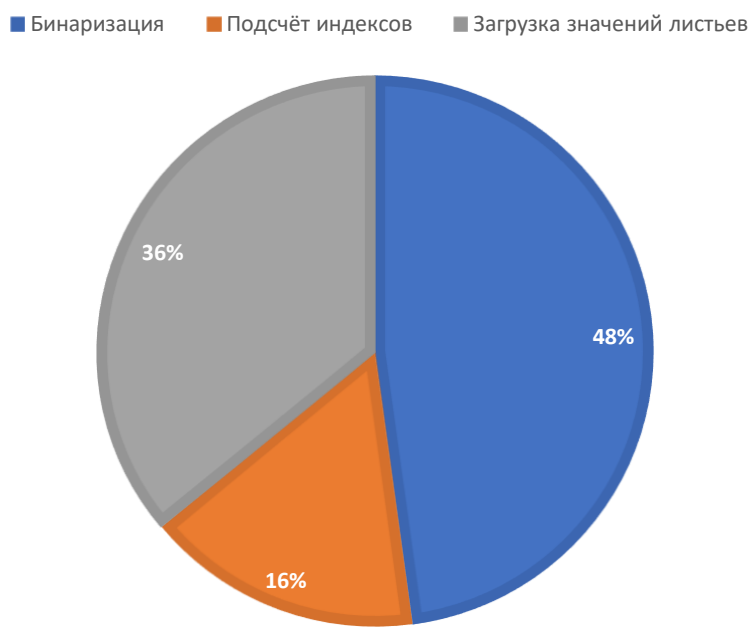

*Рисунок 3.1 Доля времени, затраченного в среднем на каждую из стадий для применения модели. Времена замерены с использованием базового алгоритма CatBoost, с батчами по 1024 объекта и равным числом запусков с транспонированной и нетранспонированной матрицей.*

## 4. Архитектура процессоров Intel

Intel с выходом каждого поколения процессоров публикует документацию [5], рассказывающую, как правильно оптимизировать ассемблерный код программ. Мы пишем реализации алгоритмов на C++ с интринсиками, и большинство возможных оптимизаций компилятор выполняет за нас. Тем не менее возможности оптимизации компилятором ограничены, и в некоторых случаях для эффективной реализации всё ещё важно понимать как процессор выполняет инструкции. Рассмотрим некоторые важные детали работы ядра.

### 4.1. Конвейер и порты

Перед тем как выполнять инструкцию, её нужно декодировать. Для этого инструкции подаются на конвейер. На конвейере инструкции обрабатываются в том порядке, в котором они записаны в коде программы. Одновременно на конвейере может находиться несколько инструкций, в зависимости от поколения. В случае, если из-за ветвлений не известно, какие инструкция должны выполняться дальше, так как это зависит от результата операций, которые ещё не выполнены, ядро пытается предсказать, по какой ветви пойдёт исполнение дальше, чтобы избежать простаивания конвейера. Предсказания делаются на основе кода и статистик, собранных за время выполнения программы. Если предсказание оказывается корректным, выполнение продолжается без проблем, если же нет, то всю проделанную работу приходится отбрасывать и начинать обрабатывать другую ветвь.

Для оптимизации работы алгоритма нам важно минимизировать число неправильных предсказаний ветви, так как такие события серьёзно сказываются на производительности. Лучший способ сделать это – минимизировать число ветвлений. Циклы можно развернуть (если число итераций константно) или векторизовать, повторив тело цикла несколько раз. У этого есть обратная сторона: это повышает нагрузку на кэш инструкций и приводит к невозможности работы LSD (Loop Stream Detector), который позволяет выполнять небольшие циклы без декодирования инструкций на каждой итерации. Однако, в общем случае Intel рекомендует разворачивать циклы.

Декодированные и подготовленные к выполнению инструкции передаются планировщику ядра. Ядро имеет несколько портов (10 на Ice Lake), каждый из которых умеет выполнять некоторые инструкции. Число портов и доступные инструкции зависят от поколения процессора. У каждого порта есть свой список поддерживаемых им инструкций. Некоторые инструкции можно выполнять только на одном из портов, для других можно выбирать один из нескольких портов. Планировщик размещает инструкции на портах. В начале каждого такта планировщик может запустить несколько инструкций, но не более одной инструкции на порт. Планировщик не обязательно отдаёт на выполнение инструкции в том же порядке, в котором они были записаны в коде. Для начала выполнения инструкции достаточно, чтобы её входные данные были готовы. Каждая инструкция выполняется на порте один или несколько тактов, но начинать выполнение инструкции на порте можно каждый такт. Таким образом даже в рамках одного порта может выполняться несколько инструкций одновременно, начатых в разное время.

Оптимизировать выполнение инструкций сложнее. Разумеется, нужно стараться минимизировать число инструкций и их латентность (число тактов на порте), но также нужно максимально утилизировать порты, выполняя как можно больше инструкций одновременно. Для этого нужно иметь несколько независимых потоков данных, чтобы соседние инструкции по возможности не зависели от результатов друг друга и могли выполняться одновременно. Так же имеет смысл чередовать инструкции разного типа. Например, логические инструкции и инструкции работы с памятью используют разные порты, поэтому они не будут конфликтовать за ресурсы в планировщике и их можно ставить рядом.

### 4.2. Память

В то время, как большинство логических операций выполняется за 1–3 такта, чтение из памяти и запись в память занимают значительно больше. Чтение незакэшированных данных может занимать десятки тактов. Для закэшированных сильно меньше, так на Ice Lake чтение из L1 кэша занимает 5 тактов, а из L2 – 13 (см. таблицу 4.1). Поэтому желательно, чтобы данные оказались в кэше ядра. При последовательном чтении процессор может загружать нужные данные автоматически, но можно использовать prefetch инструкции для принудительной загрузки данных в кэш.

Однако, не всегда возможно добиться присутствия всех нужных данных в L1 или L2 кэше. Пропускная способность кэшей сильно ограниченна. Размеры кэшей тоже. Поэтому имеет смысл уменьшит размер промежуточных вычислений, чтобы он умещался в кэш. Мы не можем повлиять на размер модели, но мы можем обрабатывать объекты блоками и выбирать размеры блоков так, чтобы результаты бинаризации влезали в кэш.

*Таблица 4.1 Параметры кэшей на Ice Lake*

| Уровень | Размер | Размер линии (байты) | Задержка (такты) | Пропускная способность (байт/такт) |
|---|---|---|---|---|
| **L1D** | 48Кб | 64 | 5 | 128 |
| **L2** | 512Кб | 64 | 13 | 48 |
| **L3** | До 2 Мб на ядро | 64 | Зависит от числа ядер | 21 |

Так же важно заметить, что мы загружаем данные из памяти кэш-линиями. Адреса этих линий выровнены. То есть вся память покрыта непересекающимися кэш-линиями. Загрузка любого числа байт из линии в регистры может занимать столько же, сколько загрузка всей линии.

### 4.3. SIMD

Для удобства разработки мы пишем код на C++, но используем интринсики для работы с SIMD регистрами и инструкциями.

Процессоры от Intel (и AMD) могут иметь разные SSE (Streaming SIMD Extensions) и AVX (Advanced Vector Extensions) расширения. В зависимости от расширений процессора нам может быть доступно разное число регистров разной длины (см. таблицу 4.2), а также разный набор инструкций. Также с AVX-512 доступно 8 специальных 64-битных opmask регистров, предназначенных для операций с битовыми масками.

Более длинные регистры позволяют нам обрабатывать больше данных параллельно и более эффективно работать с памятью. Так же AVX-512 предоставляет огромное количество новых инструкций, некоторые из которых могут быть полезны нам.

*Таблица 4.2 Число регистров в зависимости от расширений*

| Расширение | Размер регистров (биты) | Количество регистров |
|---|---|---|
| **SSE** | 128 | 16 |
| **AVX, AVX2** | 256 | 16 |
| **AVX-512VL** | 128, 256 | 32 |
| **AVX-512F** | 512 | 32 |

При написании кода необходимо помнить про число регистров и не использовать одновременно больше SSE/AVX переменных, чем у нас есть регистров, так как иначе часть из них будет писаться в память.

## 5. Методология исследования

Нами предлагается ряд гипотез о том, какие алгоритмы могут давать улучшение. Для каждой перспективной мы реализуем этот алгоритм. По возможности реализации пишутся для разных расширений процессора: SSE3, AVX2 и AVX-512.

Полученные алгоритмы применения тестируются на разном числе объектов на корректность и время работы. Для всех тестов используется процессор семейства Intel Ice Lake. Для большей точности замеры выполняются несколько тысяч раз. Итоговым временем работы считается среднее значение CPU time по всем запускам.

Все реализации компилируются с флагами с поддержкой AVX-512. Это означает, что SSE3 и AVX2 реализации на практике могут работать немного медленнее, чем показано в тестах, но с другой стороны ставит алгоритмы в более равные условия.

Мы замеряем время применения модели epsilon8k_64 [6], которая используется в бенчмарке, сравнивающем время применения CatBoost с другими библиотеками. Эта модель решает задачу бинарной классификации. Каждый объект epsilon датасета описывается 2000 числовыми признаками [7]. Модель содержит по 64 порога для каждого признака и содержит 8000 деревьев высоты 6. Данная модель не имеет категориальных признаков. Для заметки – параметры по умолчанию: 1000 деревьев, высота 6, 254 порога [9].

В качестве базовой (baseline) используется реализация CatBoost, опубликованная на GitHub.

Все разработанные алгоритмы выложены на GitHub вместе с инструкцией по запуску. Ссылка: https://github.com/alexmir1/catboost-cpu-evaluation-optimization

## 6. Бинаризация

Прежде чем начинать обрабатывать деревья, нужно выполнить бинаризацию – превратить числовые признаки в бинарные, сравнением их с порогами. Один числовой признак может превращаться в несколько бинарных, а один бинарный признак может использоваться в нескольких деревьях. Поэтому этот процесс выполняется отдельно от обхода деревьев.

Рассмотрим алгоритм бинаризации CatBoost подробнее и выявим возможности для оптимизаций.

### 6.1. Транспонирование

Модель принимает объекты батчами (пачками) произвольного размера. На вход модель получает двумерную матрицу 32-битных чисел с плавающей точкой, содержащую числовые признаки объектов батча. Эта матрица может быть нетранспонированной и иметь размер число объектов на число признаков, или транспонированной и иметь размер число признаков на число объектов. То есть, если на входе нетранспонированная матрица, то разные числовые признаки одного объекта лежат рядом в памяти, а если транспонированная, то рядом находится один числовой признак для разных объектов.

Если упростить некоторые детали, в CatBoost при бинаризации блока внешний цикл идёт по числовым признакам, внутри него цикл по объектам блока, внутри него цикл по порогам. Таким образом в рамках блока алгоритм читает входную матрицу по признакам. На первый взгляд может показаться, что такой алгоритм всегда будет работать быстрее с транспонированной матрицей на входе, так как в таком случае значения одного числового признака для разных объектов лежат близко в памяти. Однако нетранспонированную матрицу проще делить на блоки.

### 6.2. Блоки

Мы можем посчитать и записать бинаризацию для всех объектов батча сразу, но это может потребовать слишком много дополнительной памяти. Объём доступной памяти может быть ограничен, к тому же желательно, чтобы все промежуточные результаты умещались в кэш. Поэтому имеет смысл разделять батч на блоки, и обрабатывать каждый блок по отдельности.

Блок не должен быть слишком большим в силу описанных выше причин, но он так же не должен быть слишком маленьким, так как это может помешать эффективной векторизации на следующих стадиях. Базовая реализация CatBoost использует блоки размером 128 объектов. В случае с моделью с 2000 квантилями на объект, как та, что используется в тестах, на Ice Lake в L1D кэш влезут результаты бинаризации только для 24 объектов, а в L2 – для 262 объектов. И это в предположении, что в кэш пишутся только результаты бинаризации (что не так),

### 6.3. SIMD

CatBoost использует SSE инструкции для сравнения с NaN или с порогами 4 числовых признаков за инструкцию, для упаковки результатов сравнения (бинаризации) в квантили, и для записи результатов по 128 бит за инструкцию. При упаковке в квантили из 4 регистров с 4-мя 32-битными результатами сравнениями получается один регистр с 16-ю 8-битными квантилями. Если размер блока не кратен 16, то остаток обрабатывается без SSE инструкций.

Самое простое, что мы можем сделать, это использовать 256-битные или 512-битные регистры, которые доступны с AVX2 и AVX-512 расширениями. Это позволяет обрабатывать в 2 или в 4 раза больше данных за инструкцию. В таком случае если размер блока не кратен 32 или 64 соответственно, то для остатка выполняем операции без AVX инструкций.

### 6.4. Формат результата бинаризации

Результатом бинаризации является матрица однобайтовых квантилей размером число числовых признаков на размер блока. Если сравнивать с MatrixNet, где матрица состоит из бинарных признаков и имеет размер число бинарных признаков на размер блока, то видно, что матрица в CatBoost требует меньше памяти, если число порогов на один числовой признак больше 8, а их как правило 64–254 [9]. Это сильно уменьшает нагрузку на кэш. Так же этот формат даёт большую гибкость на размер блока и позволяет оптимальнее работать в случаях, когда число объектов не кратно 128 [10].

С другой стороны, формат результата бинаризации CatBoost требует больше логических операций на упаковку бинарных признаков в квантили и их распаковку, но как правило выполнение логических операций занимает сильно меньше тактов, чем работа с памятью. Такой формат также увеличивает в 8 раз объём читаемых данных на стадии вычисления индексов, так как вместо однобитных бинарных признаков приходится читать однобайтные квантили.

Однако разные деревья требуют разных бинарных признаков, поэтому чтение квантилей производится не по порядку, что усложняет преждевременную загрузку в кэш, а следовательно выигрыш от влезающего в кэш бинаризованного блока должен быть больше. И несмотря на все недостатки, вероятно формат, в котором записывается результат бинаризации CatBoost, более оптимален за счёт меньшего размера, и пробовать менять его мы не будем.

### 6.5. Бинарный поиск

Так же можно воспользоваться упорядоченностью порогов, и считать квантили не сравнением со всеми порогами, а, например, бинарным поиском. Однако, учитывая, что это даст большое число ветвлений, и вызовет связанные с этим проблемы, описанные выше, а также учитывая, что число порогов у нас невелико, скорее всего это не даст ускорения. Например, в используемой нами модели до 64 порогов на числовой признак, а значит можно посчитать квантили либо с помощью 64 векторизуемых сравнений, либо с помощью 8 невекторизуемых. AVX-512 позволяет нам сравнивать 16 чисел с порогами за инструкцию, позволяя выполнить 64 сравнения за 4 инструкции. Поэтому использовать бинарный поиск мы не будем.

### 6.6. Выводы

Данные могут поступать в транспонированном и нетранспонированном виде. Формат выбирается пользователем библиотеки, в зависимости от того, откуда берутся данные.

Для оптимизации бинаризации и квантилизации стоит попробовать увеличить размеры регистров, при наличии новых расширений. Так же стоит попробовать протестировать алгоритмы с разными размерами блоков.

Формат бинаризации мы менять не будем, бинарный поиск использовать не будем.

## 7. Подсчёт индексов

После бинаризации блока CatBoost идёт по деревьям, загружает и вычисляет нужные для очередного дерева бинарные признаки, и превращает их в индексы листа с ответом. SSE инструкции используются для загрузки квантилей для 16 объектов за инструкцию, распаковки в бинарные признаки 16-и квантилей за инструкцию, и превращения с помощью битовых операций нескольких регистров с бинарными признаками в один регистр с индексами для 16 объектов. Если размер блока не кратен 16, то для остатка все операции производятся без SSE инструкций.

Все эти операции тривиальны, единственная возможность для оптимизации здесь – это замена 128-битных регистров на 256-битные или 512-битные. В таком случае если размер блока не кратен 32 или 64 соответственно, то для остатка выполняем операции без AVX инструкций.

# 8. Вычисление индексов листьев

По полученным индексам сразу загружаются и суммируются ответы. Значения в листьях представляют из себя 64-битные числа с плавающей точкой. CatBoost загружает ответы наивно без SIMD инструкций. SSE инструкции используются для суммирования ответов из разных деревьев.

## 8.1. Gather

SSE не позволяет векторизовать загрузку по произвольным адресам. Но с AVX2 было добавлено семейство инструкций Gather для 128-битных и 256-битных регистров, которые позволяют по базовому адресу и вектору индексов загрузить вектор чисел по советующим отступам. С AVX-512 это семейство было расширено для 512-битных регистров. Мы можем использовать эти инструкции для эффективной загрузки.

## 8.2. Перестановки

Альтернативой загрузки по индексам может быть последовательная загрузка всех значений из листьев в регистры, а затем выполнение перестановки с использованием инструкций семейства Permute, чтобы сопоставить индексам ответы. Мы можем выровнять в памяти массивы с ответами деревьев и загружать их в регистры целыми кэш-линями, что значительно быстрее загрузки отдельных чисел.

В дополнение к этому, суммарный объём загружаемых данных также может быть меньше. Как правило в моделях используются деревья высоты 6, используемая нами модель не исключение. Это даёт 64 листа в дереве. А объектов в блоке может быть больше 64, и для каждого загружается по значению.

В 256-битный регистр влезает только 4 значения, а значит потребуется 16 регистров. С AVX2 у нас всего 16 регистров, а ещё нужны регистры для индексов и результатов перестановки, значит у нас не хватит регистров для такого алгоритма с AVX2.

В 512-битный регистр влезает уже 8 значений, а значит потребуется всего 8 регистров на дерево. С AVX-512 у нас 32 регистра, значит их более чем хватает.

Разберём алгоритм подробнее. Мы имеем деревья высоты 6, а значит индексы 6-битные. Получается, что индексы нужно разделить на две части: первые 3 бита задают номер регистра, последние 3 – индекс в регистре. В результате выполнения перестановки мы получаем регистр, имеющий до 8 значений. Значит мы можем обрабатывать объекты группами по 8. Мы всё ещё хотим избежать ветвлений, поэтому для каждой группы будем выполнять перестановку чисел с каждым из регистров с ответами.

Инструкция *vpermpd* может принимать до 4 аргументов: регистр со значениями для перестановки, индексы перестановки, маска, задающая по каким из индексов выполнять перестановку, и регистр со значениями по умолчанию; и возвращает один регистр, содержащий на каждой позиции либо значение из регистра с ответами, если значение маски равно 1 для этой позиции, либо значение по умолчанию с соответствующей позиции, если значение маски 0. Получить маску мы можем сравнением на равенство первых трёх бит индексов с номером регистра, для перестановки использовать последние три бита индекса, а в качестве значения по умолчанию использовать занулённый регистр или регистр, полученный с предыдущий операции *vpermpd* для этой группы.

Как можно заметить число операций велико, и растёт как $O(2^h)$, где $h$ – высота дерева. Алгоритм можно ускорить в два раза, если уменьшить высоту деревьев на единицу. Но это помешает деревьям находить сложные закономерности в данных.

### 8.3. Перестановка FP16

Мы можем уменьшить размер массива не уменьшая высоту дерева, а уменьшив размеры элементов. Вместо 64-битных чисел с плавающей точкой мы можем использовать 16-битные. Это позволит уменьшить число требуемых регистров для хранения массива в 4 раза, а объекты можно будет обрабатывать блоками по 32 объекта, вместо 8. В итоге это позволит ускорить основную часть алгоритма перестановок в 16 раз.

Но, разумеется, это скажется на качестве. Тем не менее, не везде важно иметь высокую точность. Например, в задачах классификации использование 16-битные чисел для предсказаний может не повлиять на точность предсказаний.

Перед суммированием 16-битные значения конвертируются в 32-битные, чтобы не терять в точности ещё сильнее при складывании.

Инструкции Permute для работы с 16-битными числами были добавлены только в AVX-512. С AVX2 мы можем использовать 32-битные перестановки совместно с 16-битными числами, но тогда в результате каждой перестановки мы будем получать до 8 чисел, и хотя регистров для хранения массива потребуется всего 4, операций меньше не станет, так как нужно будет ещё выполнять сдвиги по маске, чтобы получить из 32-битной перестановки 16-битные числа. Поэтому навряд ли такой алгоритм сможет обойти базовый.

# 9. Результаты

*Таблица 9.1 Замеры времени применение батча из 1024 объектов с помощью разработанных нами алгоритмов и относительное отклонение* $d = \frac{time - base\_time}{base\_time}$

| Расширение | Загрузка ответов | Размер блока | Транспонирована | | Не транспонирована | | |
|---|---|---|---|---|---|---|---|
| | | | Время (мс) | Отклонение от базового | Время (мс) | Отклонение от базового | Отклонение от трансп. |
| **SSE** | | | | | | | |
| | **Наивная** | | | | | | |
| | | 128 | 12.430 | 0% | 10.385 | 0% | -16% |
| **AVX2** | | | | | | | |
| | **Наивная** | | | | | | |
| | | 64 | 10.990 | -12% | 7.919 | -24% | -28% |
| | | 128 | 9.702 | -22% | 8.006 | -23% | -17% |
| | | 256 | 6.934 | -44% | 12.299 | +18% | +77% |
| | **Gather** | | | | | | |
| | | 128 | 10.225 | -18% | 8.289 | -20% | -19% |
| | | 256 | 7.125 | -43% | 12.192 | +17% | +71% |
| **AVX-512** | | | | | | | |
| | **Наивная** | | | | | | |
| | | 64 | 10.517 | -15% | 8.285 | -20% | -21% |
| | | 128 | 10.100 | -19% | 8.024 | -23% | -21% |
| | | 256 | 7.162 | -42% | 12.335 | +19% | +72% |
| | | 512 | 7.156 | -42% | 13.979 | +35% | +95% |
| | **Наивная FP16** | | | | | | |
| | | 128 | 9.767 | -21% | 7.926 | -24% | -19% |
| | | 256 | 6.914 | -44% | 11.950 | +15% | +73% |
| | | 512 | 6.289 | -49% | 13.077 | +26% | +108% |
| | **Gather** | | | | | | |
| | | 128 | 10.275 | -17% | 8.338 | -20% | -19% |
| | | 256 | 7.214 | -42% | 12.407 | +19% | +72% |
| | | 512 | 6.462 | -48% | 13.373 | +29% | +107% |
| | **Перестановка** | | | | | | |
| | | 128 | 12.376 | 0% | 10.642 | +2% | -14% |
| | | 256 | 9.841 | -21% | 15.038 | +45% | +53% |
| | | 512 | 9.340 | -25% | 16.315 | +57% | +75% |
| | **Перестановка FP16** | | | | | | |
| | | 128 | 7.094 | -43% | 5.210 | -50% | -27% |
| | | 256 | 4.300 | -65% | 9.361 | -10% | +118% |
| | | 512 | 3.627 | -71% | 10.436 | 0% | +188% |

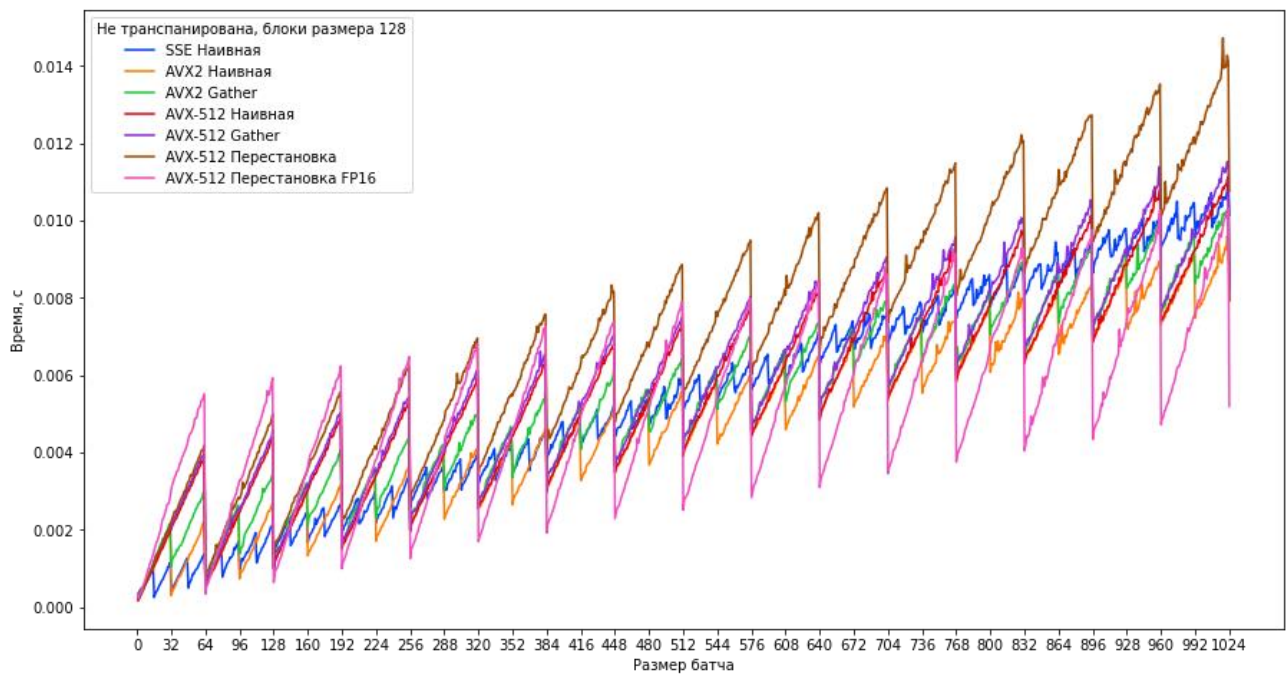

*Рисунок 9.1 График зависимости времени работы алгоритмов от числа объектов. Во всех замерах входная матрица не транспонирована, блоки по 128 объектов*

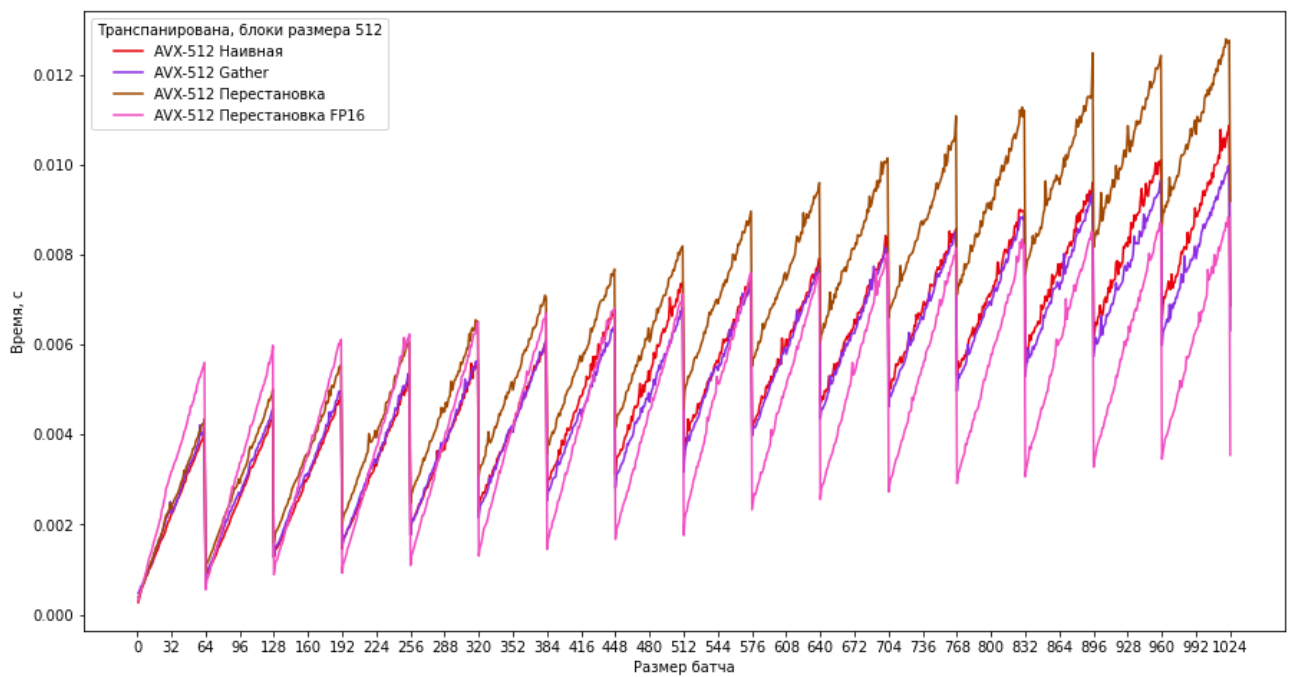

*Рисунок 9.2 График зависимости времени работы алгоритмов от числа объектов. Во всех замерах входная матрица транспонирована, блоки по 512 объектов*

## 9.1. Размеры блоков

Первое, что можно заметить, глядя на таблицу 9.1 – как отличается влияние размера блока на время между транспонированным и нетранспонированным входом. Если входная матрица не транспонирована, то видно, что минимальное время достигается при блоке размером 128 объектов. Для транспонированной матрицы видно, что чем больше объектов в блоке, тем меньше суммарное время работы.

Алгоритмы для этих двух форматов одинаковые, отличается только расположение данных в памяти. То есть различие во времени между транспонированным и нетранспонированным входом объясняется исключительно задержками, связанными с чтением из памяти. Напомним, алгоритм идёт по блокам, внутри блока по признакам, внутри признака по объектам блока.

В случае с нетранспонированной матрицей по строчкам лежат значения разных признаков одного объекта, поэтому во внутреннем цикле приходится постоянно перепрыгивать через все остальные признаки. Для каждого объекта блока приходится загружать отдельную кэш-линию. Если все эти кэш-линии будут умещаться в кэш, то при бинаризации следующего числового признака, не нужно будет их загружать заново. Поэтому имеет смысл брать не слишком большой размер блока. К тому же, как было показано выше, результаты бинаризации блока размера 128 влезают в L2 кэш, а блока размера 256 – уже нет. Слишком маленький размер блока может помешать эффективной утилизации портов процессора. Этим объясняется то, что оптимальный размер блока равен 128.

В случае с транспонированной матрицей по строчкам лежат значения одного признака для разных объектов, поэтому во внутреннем цикле данные читаются подряд, но при переходе от одного признака к другому, приходится перепрыгивать через значения признаков для всех объектов других блоков. Мы можем видеть в таблице 9.1, что при небольшом размере блока, чтение из транспонированной матрицы выполняется медленнее чтения из нетранспонированной. Это объясняется тем, что в случае с нетранспонированной матрицей, выполняется как бы параллельное чтение из нескольких массивов (строк матрицы), а в случае с транспонированной выполняется поочередное чтение массивов (отрезков строк матрицы), разбросанных по памяти. При переходе от одного такого массива к другому, есть высокий риск того, что вместо нужных кэш-линий, будут грузиться кэш-линии с признаками объектов других блоков. Чем больше размер блока, тем меньше таких переходов, и эффект, видимо, настолько сильный, что перевешивает все недостатки большого размера блока.

### 9.2. Увеличение размера регистров

По таблице 9.1 сравнивая время алгоритмов с размером блока равным 128 и наивной загрузкой мы можем оценить вклад от увеличения размера регистров, которое нам даёт AVX2 и AVX-512. 256-битные регистры уменьшают время работы примерно на 20%. 512-битные регистры не дают никакого прироста скорости по сравнению с 256-битными регистрами.

### 9.3. Алгоритмы загрузки значений листьев

В таблице 9.1 видно, что Gather не даёт значительных преимуществ перед наивной загрузкой. Векторизация не сильно помогает эффективно группировать операции чтения.

Перестановки 64-битных чисел только замедляют эту стадию.

Перестановки 16-битных чисел же дают значительную оптимизацию. Этот алгоритм опережает базовое время на 71% и 50% для транспонированной и нетранспонированной входных матриц соответственно, в то время как с наивной загрузкой мы получали опережение только на 49% и 24% соответственно. То есть алгоритм ускоряет применение на 42% и 35% соответственно. С большим размером блока ускорение сильнее, что объясняется тем, что загрузка из памяти производится один раз для всего блока.

Использование FP16 чисел также помогает ускорить и наивный алгоритм загрузки, но разница во времени далеко не такая большая, как с перестановками.

### 9.4. Зависимость времени от числа объектов

Рассмотри рисунки 9.1 и 9.2. Первое, что бросается в глаза: эффект от групп и обработки без SIMD. Напомним, в зависимости от размера регистров, объекты обрабатываются группами по 16, 32 или 64 объекта. Если число объектов не кратно 16/32/64, то остаток может обрабатываться без SIMD инструкций. Это позволяет нам не делать лишнюю работу, вызванную недозаполненными регистрами. Глядя на графики видно, что намного оптимальнее было бы игнорировать недозаполненные регистры и все равно использовать SIMD инструкции. Время обработки нескольких объектов без SIMD почти всегда занимает больше, чем обработка целой группы с SIMD инструкциями.

Также на графике 9.2 можно увидеть резкий переход на границе блока в 512 объектов. Если смотреть только на батчи, кратные 64, то видно, что с 64 до 512 рост времени замедляется, однако при переходе к новому блоку происходит резкий скачок и всё начинается сначала. Такой же эффект, но с блоками размера 128, можно разглядеть графике 9.1. Это объясняется тем, что при нетранспонированной матрице алгоритм достигает максимальной эффективности, когда размер батча кратен выбранному оптимальному размеру блока. При транспонированной матрице максимальная эффективность достигается, когда размер батча равен размеру блока.

### 9.5. Потеря точности с FP16

Как мы писали выше, FP16 может уменьшать качество предсказаний. Проверим насколько. Напомним, модель решает задачу бинарной классификации. Мы выяснили, что на всём epsilon датасете предсказанные классы моделью с 64-битными значениями в листьях полностью совпали с классами, предсказанными моделью с 16-битными значениями в листьях. То есть потери точности нет.

Но мы можем также посмотреть на уверенность модели в предсказаниях. Модель сравнивает сумму ответов деревьев с нулём, чтобы получить предсказание принадлежности к классу. Результаты сравнения этих сумм приведены в таблице 9.2.

*Таблица 9.2 Значения метрик отклонения предсказаний модели с 16-битными значениями в листьях от модели с 64-битными*

| Метрика | Значение |
|---|---|
| Максимальное отклонение | 3.14303E-04 |
| Среднее отклонение | 5.28827E-05 |
| Медианное отклонение | 4.43792E-05 |
| Корень из среднего квадрата отклонения | 6.65985E-05 |

Как можно видеть, не только предсказания принадлежности классам совпадают, но и предсказанные числа довольно близки. А потому такой алгоритм отлично подходит для задач классификации, и хорошо для задач регрессии.

## 10. Ограничения и дальнейшая работа

### 10.1. Возможные направления дальнейших исследований

Мы не стали проверять некоторые гипотезы, так как скорее всего они не окажутся успешными, но тем не менее их можно проверить. Можно попробовать поменять формат результатов бинаризации. Можно попробовать использовать бинарный поиск для бинаризации. Можно попробовать сделать 16-битные перестановки с AVX2.

Текущие реализации загрузки значений из листьев не работают, если используется многоклассовая классификация. Можно попробовать сделать реализацию перестановок FP16 для этого случая. Если в каждом из листьев по 2 или по 4 значения, то можно комбинировать 16-битные числа в 32-х или 64-битные и переставлять их.

Также, можно попробовать высчитывать индексы с меньшим размером блока, чем используется для бинаризации. Если для вычисления индексов оптимальный размер блока 128 (так как квантили 128 объектов влезают в кэш), а для бинаризации – 512, то можно выполнять бинаризацию на блоке из 512 объектов, но записывать квантили в 4 разных массива для разных подблоков, и выполнять оставшиеся стадии на этих подблоках.

Мы можем попробовать и другие подходы, ухудшающие качество. Например, можно принимать на вход матрицу 16-битных чисел, а не 32-битных. Это должно сильно ускорить бинаризацию. Особенно быстро должно работать с расширением AVX-512FP16, которое было добавлено с новым поколением Alder Lake, и которое позволяет эффективно сравнивать 16-битные числа с плавающей точкой. Без этого расширения эффективно можно только конвертировать эти числа в 32 или 64-битные.

В этой работе мы не исследовали оптимизации для категориальных признаков, так как one-hot кодирование редко используется при числе категорий больше двух [8], а ctr конвертацию навряд ли можно векторизовать. Тем не менее, можно попробовать найти другие подходы для оптимизации ctr.

Наконец, мы можем использовать многопоточность. Можно легко разделить батч на потоки, и выполнять их на разных ядрах. Это повысит пропускную способность, но не уменьшит время обработки небольшого числа объектов (latency). Уменьшить latency можно делением на потоки признаков на стадии бинаризации и деревьев на следующих стадиях.

### 10.2. Внедрение

CatBoost работает не только на Intel и AMD процессорах, но и на ARM архитектуре на мобильных устройствах. При компиляции с помощью специальной библиотеки SSE инструкции заменяются на аналогичные ARM Neon инструкции. Но AVX инструкции заменять не на что: Neon поддерживает только 128-битные регистры. Существует расширение ARM SVX (Scalable Vector Extension), которое позволяет в процессе выполнения кода масштабировать операции в зависимости от размера доступных регистров. Однако на момент написание статьи это расширение не распространено и имплементировано только на некоторых процессорах, предназначенных для суперкомпьютеров. В связи с этим SSE реализацию придётся оставить или сделать отдельную реализацию под Xeon.

На используемых в персональных компьютерах процессорах как правило поддерживается AVX2, но AVX-512 встречается пока редко. Неизвестно, насколько будет распространён AVX-512 на них в будущем, так как на новом поколение процессоров Alder Lake при включении Efficiency ядер AVX-512 не доступен, несмотря на то что Performance ядра его поддерживают. На используемых в серверах процессорах как правило AVX-512 доступен.

Можно использовать разные реализации с разными параметрами в зависимости от наличия расширений процессора и размера кэша, но это потребует сильно увеличить кодовую базу, что может усложнить дальнейшую разработку и тестирование. Поэтому внедрение любого из алгоритмов будет требовать компромиссов.

## 11. Литература и ссылки

4. GitHub CatBoost репозиторий: https://github.com/catboost/catboost (дата обращения 17.05.2022)

5. 2022 Intel® 64 and IA-32 Architectures Optimization Reference Manual https://cdrdv2.intel.com/v1/dl/getContent/671488 (дата обращения 17.05.2022)

6. Epsilon8k_64 на GitHub https://github.com/catboost/catboost/tree/master/catboost/benchmarks/model_evaluation_speed (дата обращения 17.05.2022)

7. Epsilon dataset https://catboost.ai/en/docs/concepts/python-reference_datasets_epsilon (дата обращения 17.05.2022)

8. Максимальное число классов, при котором допускается one-hot кодирование, по умолчанию https://catboost.ai/en/docs/references/training-parameters/common#one_hot_max_size (дата обращения 17.05.2022)

9. Параметры обучения по умолчанию https://catboost.ai/en/docs/references/training-parameters/ (дата обращения 17.05.2022)

10. Миронов Алексей, Ильнур Хузиев. 2021. Оптимизации применения решающих деревьев с помощью SIMD инструкций https://arxiv.org/abs/2205.07307
Все разработанные алгоритмы выложены на GitHub вместе с инструкцией по запуску. Ссылка: https://github.com/alexmir1/catboost-cpu-evaluation-optimization

29